\documentclass{IEEEtran}  % Comment this line out if you need a4paper

\usepackage[utf8]{inputenc} 
\PassOptionsToPackage{hyphens}{url}\usepackage{hyperref}

\usepackage{amsmath, mathtools, amssymb}
\usepackage{multirow, multicol}
\usepackage{graphicx}
\usepackage{caption}
\usepackage{subcaption}
\usepackage{color, soul}
\usepackage{verbatim}
\usepackage{hyperref}
\usepackage{float}
\usepackage{cite}
\usepackage{tabularx}
\usepackage{makecell}
\usepackage{todonotes}
\hypersetup{
  colorlinks   = true, %Colors links instead of ugly boxes
  urlcolor     = blue, %Color for external hyperlinks
  linkcolor    = blue, %Color of internal links
  citecolor   = red %Color of citations
}

\title{\LARGE \bf
Improving Needle Penetration via Precise Rotational Insertion Using Iterative Learning Control
}
\author{Yasamin Foroutani\textsuperscript{1}, Yasamin Mousavi-Motlagh \textsuperscript{2}, Aya Barzelay \textsuperscript{2}, Tsu-Chin Tsao\textsuperscript{1}
\\ \textsuperscript{1} Dept. of Mechanical and Aerospace Engineering, University of California, Los Angeles \\ \textsuperscript{2} Jules Stein Eye Institute, University of California, Los Angeles }

\begin{document}
\maketitle

%%%%%%%%%%%%%%%%%%%%%%%%%%%%%%%%%%%%%%%%%%%%%%%%%%%%%%%%%%%%%%%%%%%%%%%%%%%%%%%%
% Abstract 
%%%%%%%%%%%%%%%%%%%%%%%%%%%%%%%%%%%%%%%%%%%%%%%%%%%%%%%%%%%%%%%%%%%%%%%%%%%%%%%%
\begin{abstract}
Achieving precise control of robotic tool paths is often challenged by inherent system misalignments, unmodeled dynamics, and actuation inaccuracies. This work introduces an Iterative Learning Control (ILC) strategy to enable precise rotational insertion of a tool during robotic surgery, improving penetration efficacy and safety compared to straight insertion tested in subretinal injection. A 4 degree of freedom (DOF) robot manipulator is used, where misalignment of the fourth joint complicates the simple application of needle rotation, motivating an ILC approach that iteratively adjusts joint commands based on positional feedback. The process begins with calibrating the forward kinematics for the chosen surgical tool to achieve higher accuracy, followed by successive ILC iterations guided by Optical Coherence Tomography (OCT) volume scans to measure the error and refine control inputs. Experimental results, tested on subretinal injection tasks on ex vivo pig eyes, show that the optimized trajectory resulted in higher success rates in tissue penetration and subretinal injection compared to straight insertion, demonstrating the effectiveness of ILC in overcoming misalignment challenges. This approach offers potential applications for other high precision robot tasks requiring controlled insertions as well.
\end{abstract}

\section{Introduction}
\begin{figure}
    \centering
    \includegraphics[width=\linewidth]{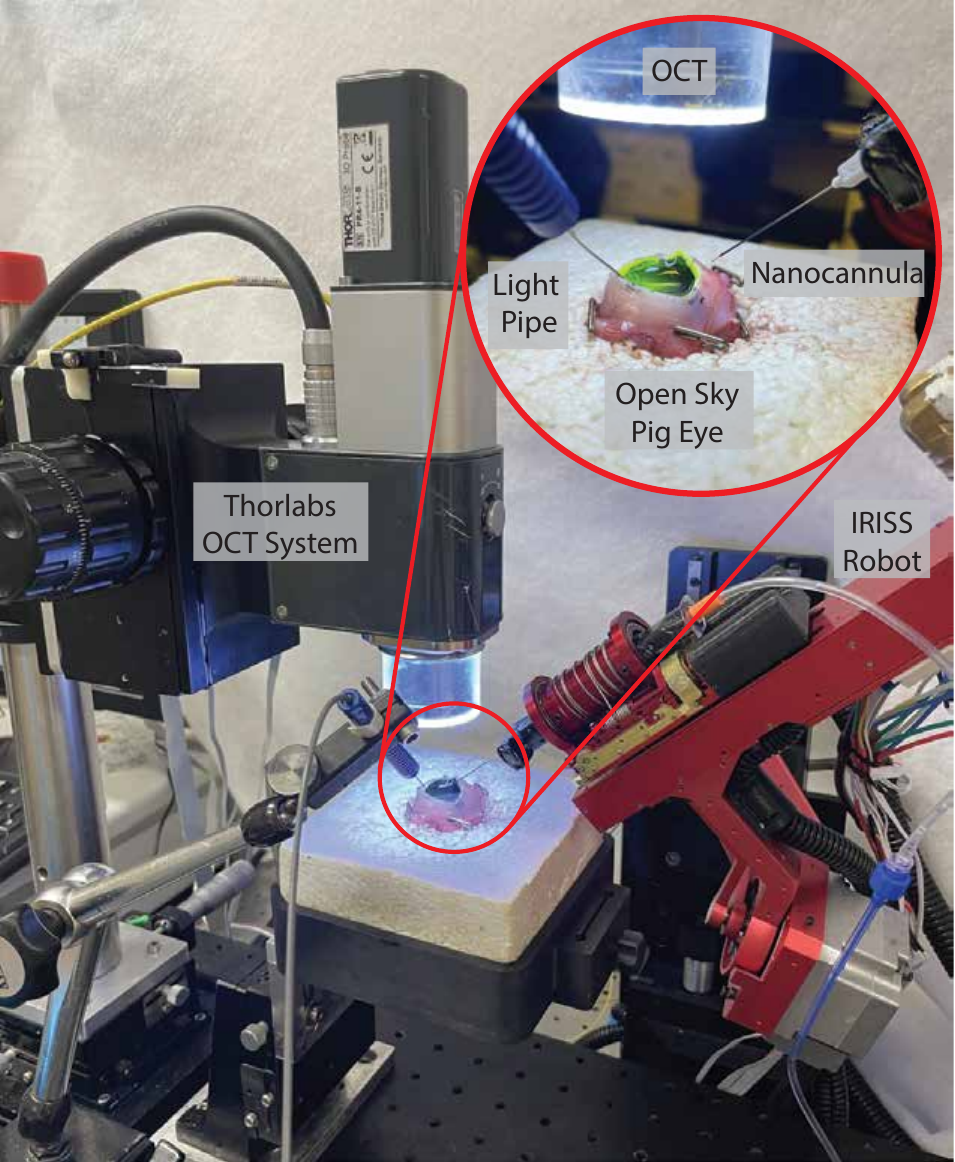}
    \caption{System overview showing the experimental setup, using Thorlabs OCT, IRISS robot, and an open sky pig eye to perform injection.}
    \label{fig:robot_oct_sys}
\end{figure}

Accurate and precise control of movement is fundamental to many scientific fields \cite{iwasaki2012high}, but it becomes even more critical in surgical applications where even minor deviations can significantly impact outcomes. Surgical procedures often demand sub-millimeter accuracy, especially in areas involving delicate tissues and confined spaces, such as ophthalmology. However, consistently achieving this level of precision can be challenging due to the inherent limitations of human motor skills, such as involuntary tremors and fatigue \cite{singh2002physiological}. These limitations are amplified in intraocular microsurgery, requiring not only steady hands, but also enhanced sensory feedback and hand-eye coordination. As a result, there has been increasing interest in augmenting surgical precision through the use of robotic systems, which can assist surgeons and automate difficult tasks, leading to improved accuracy and stability compared to manual techniques \cite{iordachita2022robotic}.\\
One critical surgical maneuver is precise needle insertion through sensitive tissues, where maintaining exact trajectory control is paramount. Consistent penetration of the target tissue using straightforward linear insertion alone is challenging due to variations in tissue resistance, elasticity differences, changes in needle sharpness, and needle deflection \cite{tsumura2019needle}. This frequently leads to tissue deformation without penetration, or sudden bounce back of the tissue following needle insertion, increasing risks of damage and overshoot \cite{pannek2024exploring}. Studies indicate that bidirectional rotational motion during needle insertion can significantly reduce contact forces and improve penetration efficacy \cite{hochman2001vitro, abolhassani2006control}, while reducing needle deflection and improving precision \cite{tsumura2016histological, lin2020simultaneously}. However, manually achieving such precise rotational motions is nearly impossible due to human motor limitations. As such, robotic systems are crucial for the development of such a surgical strategy.\\
Despite the advantages robots bring to surgery, achieving precise rotational control remains problematic. Robotic systems are susceptible to system misalignments \cite{millington2019innovative, wilson2018intraocular, kim2015effects, beasley2004kinematic, bai2021error}, unmodeled dynamics \cite{huang2024operational, cursi2020model}, and actuation inaccuracies \cite{ogihara2024intraoperative, omisore2020motion}, all of which can hinder precise tasks such as needle insertion. These issues can lead to unintentional tooltip displacements, introducing disturbances that are difficult to model and compensate for in real time \cite{abolhassani2006control, li2018needle}. Such limitations necessitate advanced control strategies to refine robot motion and maintain surgical precision throughout the operation.\\
To address the challenges of achieving precise robot control, various advanced control strategies have been developed. Traditional PID controllers remain a foundational approach in robotics due to their simplicity and ease of implementation. They fall short, however, in situations where kinematic and dynamic errors are present. Classical computed torque or inverse dynamics methods address some of these issues by correcting for nonlinearities \cite{spong2020robot}. They still depend heavily on accurate dynamic models of the robot, though, and performance degrades in the presence of uncertainties. Adaptive control strategies build on this approach by continuously adjusting control parameters to correct for the effects of system variations and model uncertainties. Such techniques ensure the system can compensate for changing dynamics during operation \cite{wei2018adaptive, kaneko1997repetitive}. These methods, while robust, can introduce complex adaptation laws that require careful tuning and significant computational resources, making implementation in real-time surgical environments challenging.\\
A common challenge shared by all the mentioned feedback approaches is the need for rigorous stability analysis and potential redesigns of the controller to account for changing dynamics and kinematic uncertainties. This often complicates integration into existing platforms or new applications. In contrast, feedforward control methods offer an attractive alternative. By proactively predicting and compensating for errors, feedforward controllers minimize the need for real-time feedback. This makes them simple to integrate, leading to enhanced performance without altering the baseline low-level controller, and has motivated complementary approaches that tune feedforward cost functions from offline optimal solutions \cite{abtahi2023automatic}.\\
For repetitive tasks where consistent trajectories are crucial, Iterative Learning Control (ILC) is a particularly effective feedforward strategy. ILC refines the control inputs by learning from previous executions, iteratively reducing errors over successive trials. This method has successfully been implemented in industrial applications such as machining \cite{kim1996iterative}, additive manufacturing \cite{hagqvist2015resistance}, flatbed printers \cite{blanken2016design}, wafer stages \cite{dijkstra2002convergence}, and industrial robot arms \cite{lee2021industrial}. 
The iterative nature of ILC makes it an ideal solution for surgical procedures such as needle insertions. It can iteratively correct for kinematic errors and misalignments, while also compensating for dynamic variations in the robot. Unlike traditional methods, ILC achieves this without the need for continuous recalculation or detailed dynamic modeling, making it an efficient and robust solution for enhancing accuracy in repetitive surgical procedures. When the dynamic model of the system is accurately known, model-based ILC approaches can be used, such as in \cite{longman2000iterative, norrlof2002time, gunnarsson2001design, tao2021robust, wang2015nonparametric}. However, in the absence of such models, data-driven ILC, using data from previous trials, is used for learning \cite{lee2022data}. \\
This work aims to implement precise rotational needle insertion using a dual-loop iterative learning control method, similar to that proposed by \cite{lee2021industrial}, to enhance injection precision and repeatability in ophthalmic surgery. In this approach, the first loop focuses on minimizing end-effector deviations by iteratively adjusting the surgical tool's trajectory, while the second loop compensates for motor dynamics to achieve accurate trajectory tracking. A key advantage of utilizing this ILC strategy is its data-driven nature, leveraging 3D Optical Coherence Tomography (OCT) imaging feedback to measure positional errors at each iteration. By continuously updating the trajectory based on this visual information, the system effectively corrects for unmodeled dynamics and tool misalignments without requiring explicit recalculation of robot parameters. This two-step architecture not only improves rotational insertion accuracy over successive trials but also provides a scalable and adaptable pipeline applicable to a wide range of other repetitive surgical tasks.\\

\section{Methodology}
\begin{figure*}
    \centering
    \includegraphics[width=0.8\textwidth]{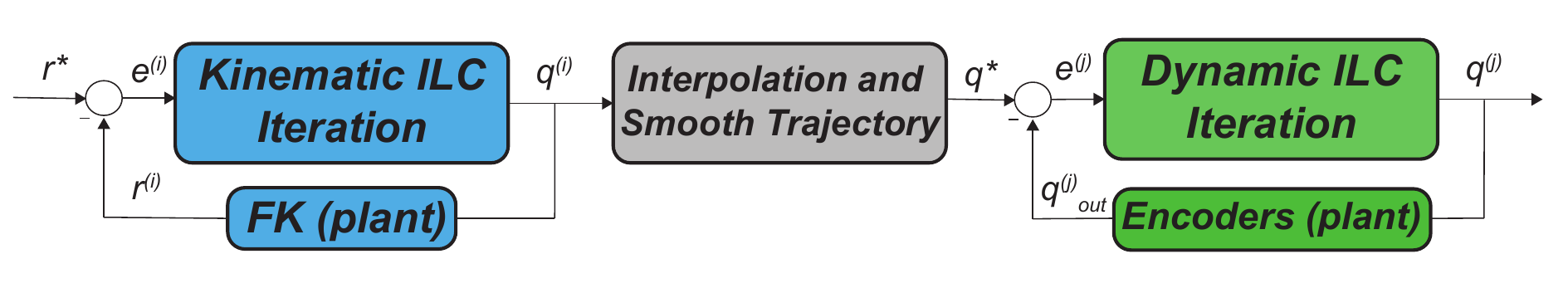}
    \caption{Learning structure for the two step ILC workflow. The kinematic ILC updates the joint values to minimize tooltip position error, and the dynamic ILC updates the derived smooth trajectory to ensure correct tracking.}
    \label{fig:ILC_blockdiagram}
\end{figure*}
\subsection{Robotic System}
In this project, we make use of the Intraocular Robotic Intervention Surgical System (IRISS), previously developed by our team \cite{wilson2018intraocular}. The IRISS is a 4 Degree of Freedom (DOF) robot, with three revolute joints and one prismatic joint, all intersecting at a remote center of motion (RCM). The robot has successfully been used to perform cataract \cite{chen2018intraocular, lee2023accurate, gerber2021robotic, chen2019semiautomated} and retinal \cite{barzelay2024transforming, gerber2020automated} surgery on phantoms and pig eyes, and as such, has proven to have the required resolution and precision to perform intraocular microsurgery. \\
To provide sensing and feedback to the system, a transpupillary spectral domain (SD) Optical Coherence Tomography (OCT) system is used. The engine provides a central wavelength of 1060 nm (Telesto II, Model No. 1060LR, Thorlabs) and contains two galvanometers with an objective lens (LSM04BB, ThorLabs) that is capable of capturing three-dimensional volumetric images (volume scans) of the target. A calibration process is done in the beginning of each experiment to find the registration between OCT and robot coordinate frames, as well as calibrating the robot's DH parameters to improve tool localization. The calibration and registration pipeline is explained in detail in \cite{lai2025safe}. The surgical setup, with the robot and OCT combination, can be seen in Fig. \ref{fig:robot_oct_sys}.\\

\subsection{Iterative Learning Control}
The proposed ILC framework is developed to address two distinct challenges in high-precision robotic tasks: correcting kinematic errors and ensuring accurate trajectory execution under dynamic conditions. The architecture can be seen in Fig. \ref{fig:ILC_blockdiagram}. Unlike the method proposed in \cite{lee2021industrial}, which used a nested loop approach, this framework separates these tasks into two sequential steps, each targeting a specific source of error.\\
In the first step, the focus is on correcting kinematic inaccuracies that result from errors in the robot's nominal kinematic values or calibration mismatches. The desired end-effector path $r^*(t)$ in task space is iteratively refined to account for these deviations. Volumetric scans acquired through Optical Coherence Tomography (OCT) are used to measure the deviation of the actual end effector position from the intended location. This information is then used to adjust the joint space trajectory over successive iterations, ensuring that the computed joint values $q^*(t)$ align the robot's tooltip position with the reference value. This step optimizes the joint space trajectory without requiring explicit system recalibration, addressing kinematic misalignments.\\
Once the optimized joint values are determined for discrete points, they are interpolated to achieve a smooth trajectory with the desired speed. At this point, the second step is performed, focusing on accurately executing the trajectory under the influence of dynamic effects, such as motor dynamics and system nonlinearities. At this stage, the updated joint space trajectory $q^*(t)$ serves as the reference, and the control inputs are iteratively refined to minimize tracking errors. Using encoder feedback, the system learns to compensate for dynamic discrepancies, ensuring precise tracking even at high speeds. This step makes use of the repetitive nature of the task to refine the input in the absence of an accurate dynamic model.\\
By structuring these steps sequentially, the framework avoids complexities involving simultaneous learning and ensures that modifications to one step do not interfere with the other. The first step provides a well-calibrated trajectory that minimizes kinematic errors, while the second step ensures precise execution of this trajectory under real-world conditions by modifying the input commands. The data-driven nature of the framework ensures adaptability to any system, enabling iterative improvements without explicit system models.\\

\subsection{Compensating Kinematic Errors} \label{subsec:ILC_kinematics}
\begin{figure}
    \centering
    \includegraphics[width=\linewidth]{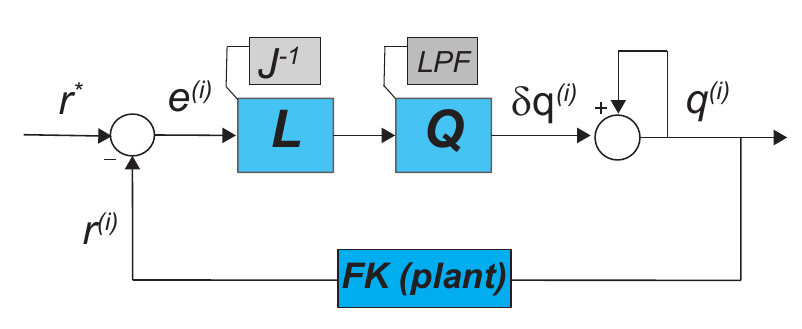}
    \caption{Initial ILC loop for correcting kinematic errors.}
    \label{fig:ILC_kinematic}
\end{figure}

Having established a sequential framework for the control, the first step focuses on addressing kinematic inaccuracies that arise from calibration errors, unmodeled offsets, or misalignments in the robotic system. These errors still exist even when joint level tracking is accurate, as the source is discrepancies between the robot's nominal forward kinematics and its actual values. To resolve this issue, an ILC approach is implemented, using 3D imaging as feedback to iteratively update the joint space trajectory and minimize end effector deviation.\\
In this step, the deviation of the end effector path compared to the desired task space trajectory $r^*(t)\in \mathbb{R}^m$ is measured, and reduced by refining joint space inputs. Note that $m$ represents the number of task space dimensions of interest (e.g. 3 cartesian coordinates of the end effector if only controlling position, and 6 if the tool orientation is also accounted for). Let the task space deviation at ILC iteration $i$ be $e_r^{(i)}(t)$, defined as the difference between the desired end effector position $r^*(t)$ and measured position $r^{(i)}(t)$:\\
\begin{equation}
    e_r^{(i)}(t) = r^*(t) - r^{(i)}(t)
\end{equation}
To minimize this deviation, the joint space correction $\delta q^{(i+1)}$ is computed iteratively:\\
\begin{equation} \label{eq:ILC_kinematic_dq_update}
    \delta q^{(i+1)} = L^{(i)} . e_r^{(i)}(t),
\end{equation}
where $L^{(i)}\in \mathbb{R}^{m\times n}$ is the learning gain matrix, which maps the $m$ dimensional task space errors to the $n$ dimensional joint space correction. The corrected joint trajectory for the next iteration is then:\\
\begin{equation} \label{eq:ILC_kinematic_q_update}
    q^{(i+1)}(t) = q^{(i)}(t) + \delta q^{(i+1)}.
\end{equation}
Here $q^{(i)}(t)$ represents the joint values at iteration $i$, and $q^{(i+1)}(t)$ is the updated trajectory designed to reduce end effector errors.\\
To find a suitable learning matrix, we will make use of the robot's forward kinematics, which relate joint space inputs $q(t)\in \mathbb{R}^n$ to task space outputs $r(t)\in\mathbb{R}^m$:
\begin{equation}
    r(t) = FK(q(t))
\end{equation}
with $FK(\cdot)$ representing the kinematic mapping for a given orientation. The Jacobian matrix $J(q) = \frac{\partial FK}{\partial q}\in\mathbb{R}^{m\times n}$ approximates this relationship locally, and can be derived analytically from the forward kinematics or numerically (see Sec. \ref{subsec:experiment_ILC1}). Its inverse $J^{-1}(q)$ can then be used to compute joint space corrections:
\begin{equation} \label{eq:ILC_kinematic_dq_from_Jinv}
    \delta q^{(i+1)} = \alpha \cdot J^{-1}\cdot e_r^{(i)}(t),
\end{equation}
where $\alpha\in(0,1]$ is a scaling factor to improve robustness when the kinematic approximation has significant errors. If the jacobian is poorly conditioned or unknown, an approximate inverse or pseudoinverse can be used.\\
In practice, the end effector deviation $e_r^{(i)}(t)$ can be measured using image feedback, in this case, OCT scans, and then mapped from the imaging frame to the robot coordinate system.\\
Using the parameters derived above, the first ILC loop can be visualized as in Fig. \ref{fig:ILC_kinematic}. This update process converges if the learning gain matrix $L$ satisfies:
\begin{equation}
    \lVert I - LJ^*\rVert <1,
\end{equation}
with $J^*$ being the actual Jacobian, and $L = \alpha J^{-1}$ is a scaled inverse of the nominal Jacobian derived from approximate kinematics. The scalar $\alpha$ can be chosen sufficiently small to mitigate errors in the Jacobian approximation and ensure convergence.\\

\begin{figure}
    \centering
    \includegraphics[width=\linewidth]{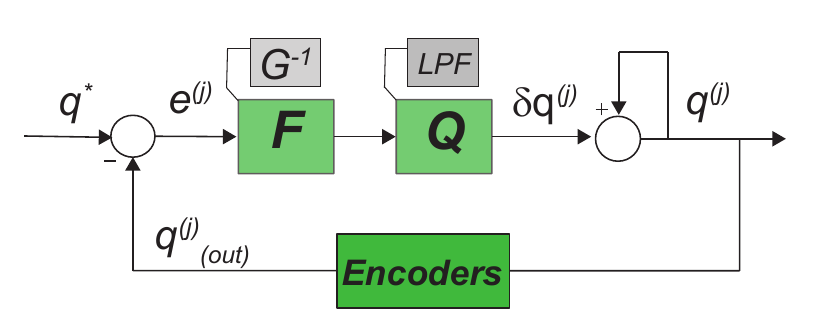}
    \caption{Second ILC loop for correcting tracking errors due to joint dynamics. }
    \label{fig:ILC_dynamics}
\end{figure}

\subsection{Compensating Motor Dynamic} \label{subsec:ILC_dynamics}
The method explained in Sec. \ref{subsec:ILC_kinematics} results in a set of joint values that achieve our desired target position at each increment. The final motion, however, must be interpolated and performed continuously to execute a smooth trajectory. This introduces challenges, especially during fast movements where the robot's dynamic effects, such as inertia, damping, and actuator limits, play a significant role in trajectory tracking. As such, additional modifications are needed to ensure precise reference tracking under these conditions.\\
To address this, the second stage of the ILC framework compensates for the dynamics of each joint by designing a strategy to update input joint values. The process begins with identifying each joint's transfer function $G(z)$, which models the relationship between the input $u(k)$, and output $q(k)$. This model includes the effects of motor dynamics, gear reduction, and other physical properties of the system. The identified transfer function can then be inverted to design the learning gain $F(z)$, to be used in the ILC loop.\\
Several strategies can be used for the plant inversion, including full model inversion \cite{chang2012high}, phase-only inversion, specialized controllers such as Zero Phase Error Tracking Controllers (ZPETC) \cite{tsao1987adaptive} and Zero Magnitude Error Tracking Controllers (ZMETC) \cite{rigney2009nonminimum}, and iterative learning-based methods such as Model-less Inversion-based Iterative Control (MIIC) \cite{kim2008model} or frequency-domain iterative approaches \cite{teng2015comparison}. Full model inversion attempts to cancel both phase and magnitude distortions, but it requires an extremely accurate model and can be highly sensitive to noise, modeling errors, and non-minimum phase dynamics. In systems with non-minimum phase behavior, exact inversion may be infeasible or destabilizing. Phase-only inversion, by contrast, focuses on correcting the dominant phase-related errors while allowing other distortions to be corrected over time through iterative learning. Data-driven approaches offer another option by progressively improving input signals based on previous error measurements without requiring an explicit model. The choice of inversion method ultimately depends on the accuracy of the plant model, the stability characteristics of the system, sensitivity to noise, and the trade-off between immediate accuracy and robustness.\\
As seen in the controller visualization in Fig. \ref{fig:ILC_dynamics}, the control law for the second stage can be expressed as:
\begin{equation} \label{eq:ILC_dynamics_update}
    u^{(j+1)}(k) = Q(z)\cdot (u^{(j)}+F(z)\cdot e^{(j)}(k),
\end{equation}
where $u^{(j)}(k)$ is the control input value at the $j-$th iteration, $e^{(j)}(k)$ is the joint error compared to the desired joint trajectory, $F(z)$ is the learning gain, and $Q(z)$ is a zero phase low pass filter. Assuming position controlled actuators, the control input is equivalent to the joint command values, resulting in $e^{(j)}(k) = q^*(k)-q^{(j)}_{out}(k)$\\
It should be noted that this process is performed on each joint and assumes that each joint can be treated as a separate single input, single output (SISO) subsystem. This assumption is generally valid because the high gear reduction in most robots minimizes coupling between joints, simplifying both analysis and controller design. Nevertheless, it is important to consider the whole robot as a multi-input multi-output (MIMO) system if coupling effects cannot be neglected.\\

\section{Experimental Setup and Results}
\subsection{Problem Statement}
\begin{figure*}
    \centering
    \includegraphics[width = 0.8\linewidth]{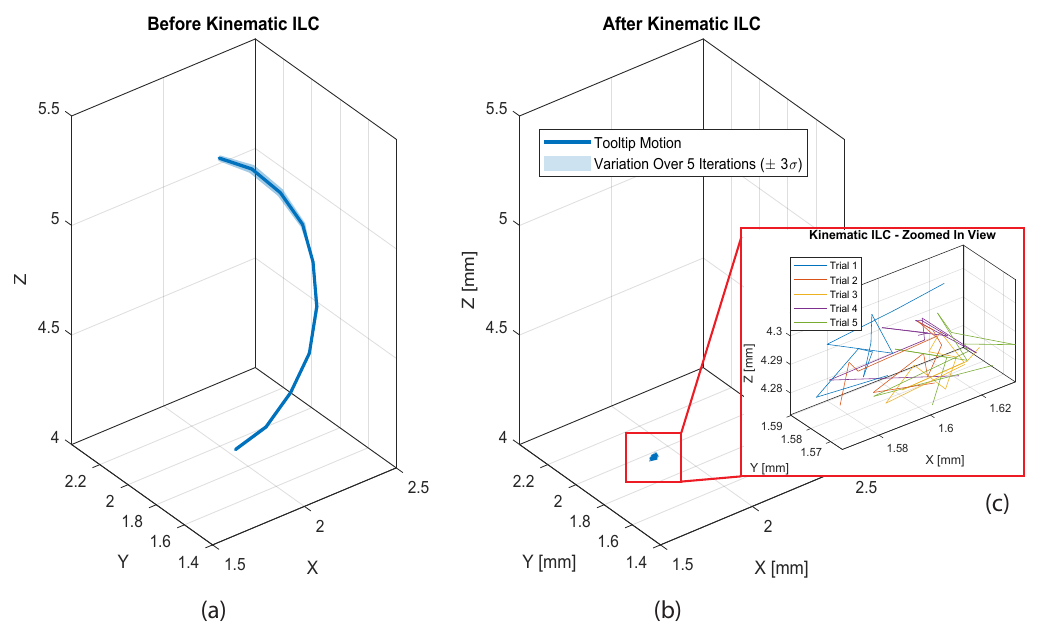}
    \caption{(a) Tooltip path before ILC (b) Tooltip path after ILC (c) Zoomed in view of tooltip path after ILC. All steps were performed 5 times to show repeatability and deviation.}
    \label{fig:kinematic_ILC_pos}
\end{figure*}

In delicate ophthalmic procedures such as subretinal injection and vein cannulation, precise needle insertion with minimal tissue damage is critical. Direct needle insertion often results in inconsistent penetration and increased needle deformation, reducing precision. Studies indicate that bidirectional rotation (oscillatory back-and-forth motion) during insertion significantly reduces tissue deformation, penetration forces, and tissue winding compared to linear or single-direction rotational insertion, which can result in tissue trauma \cite{abolhassani2006control, tsumura2016histological}. Such sinusoidal, oscillatory rotation, however, is practically impossible to perform accurately by hand due to limitations of human motor control. Our objective is to implement robotic sinusoidal oscillatory rotation around the needle's longitudinal axis, precisely maintaining needle tip location while progressing the tool, to enhance penetration efficiency, minimize tissue deformation, and eliminate tissue winding, overcoming current robotic limitations for consistent and safe subretinal injection.\\
The IRISS system is designed to facilitate such movements, with one degree of freedom dedicated to rotation about the tool's axis. In theory, this should allow the robot to perform the desired motion directly. However, in practice, a range of kinematic and dynamic factors such as calibration mismatch, needle bending, and tool misalignment, can introduce significant errors in the rotational motion of the needle. Instead of producing pure rotation about the needle’s centerline, these factors cause the tooltip to move along an elliptical trajectory. The tooltip's path for one such experiment can be seen in Fig. \ref{fig:kinematic_ILC_pos}a, where the tooltip oscillations reach approximately $850\mu m$ during multiple $180^\circ$ rotations. While this error may be negligible in many robotic tasks, it is critical when it comes to retinal surgery, where the tooltip must operate within the $\approx 300\mu m$ thick retina with up to $5\mu m$ precision. Errors of this magnitude, particularly in the Z direction, can cause the tooltip to deviate from the desired insertion point, severely damaging the delicate retinal tissue or resulting in incorrect drug delivery.\\
To address these challenges, it is necessary to develop a strategy that adjusts the motion of the other three degrees of freedom of the robot to counteract the positioning errors caused by joint inaccuracies and tool misalignment. By updating the trajectory, the tooltip can be constrained to its desired position during rotational motion. This compensation ensures that the needle rotates about its centerline while maintaining a stable tooltip position, enabling safe and precise injection.\\
Given the repetitive nature of the motion, a continuous back-and-forth rotation about the needle's axis, Iterative Learning Control (ILC) provides an ideal framework for addressing this problem. By making use of the repetitive task, it is possible to iteratively refine the control inputs, using feedback from previous iterations to reduce errors in subsequent motions. This approach, which is done offline prior to the beginning of the surgical procedure, allows the system to adapt to each individual tool used during a separate surgery. After a few iterations, the ILC algorithm converges, producing an optimized trajectory that can be easily integrated with any desired motion throughout the surgery, enabling precise tool rotation during insertion.\\

\begin{figure}
    \centering
    \includegraphics[width=0.4\textwidth]{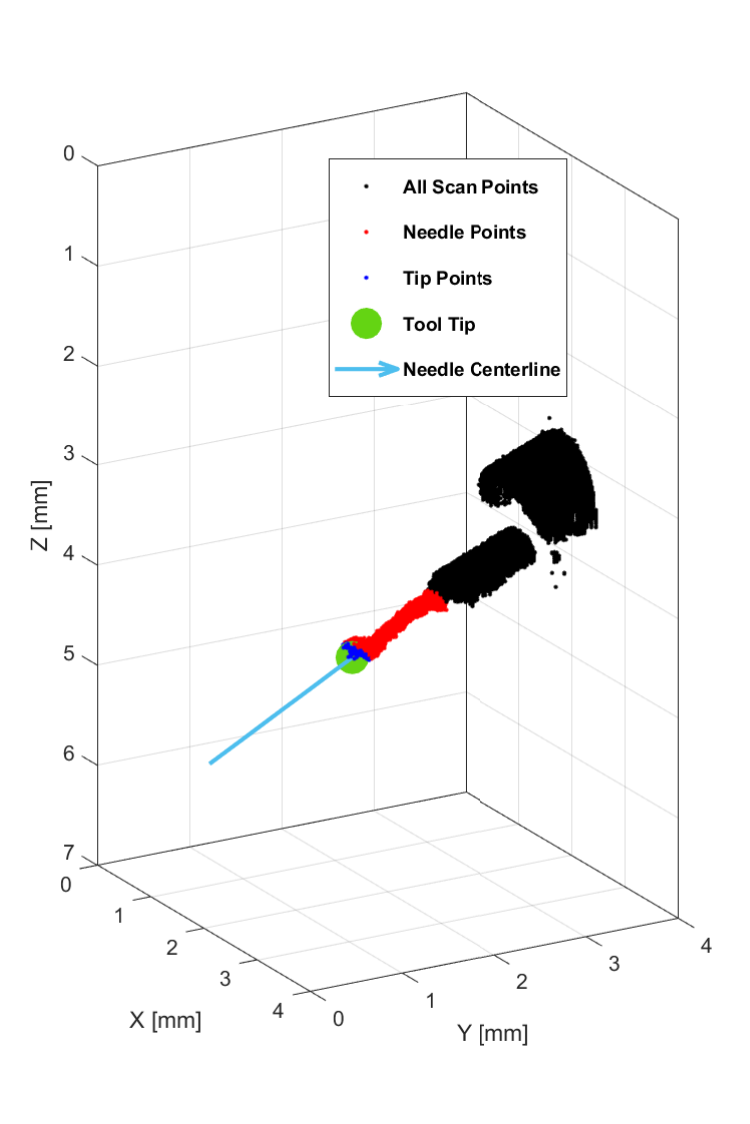}
    \caption{Segmentation and tooltip detection in a volume scan of a nanocannula used for subretinal injection}
    \label{fig:tool_segmentation}
\end{figure}

\subsection{Controller Design}
\subsubsection{Kinematic Loop Implementation} \label{subsec:experiment_ILC1}
\begin{figure}
    \centering
    \includegraphics[width=\linewidth]{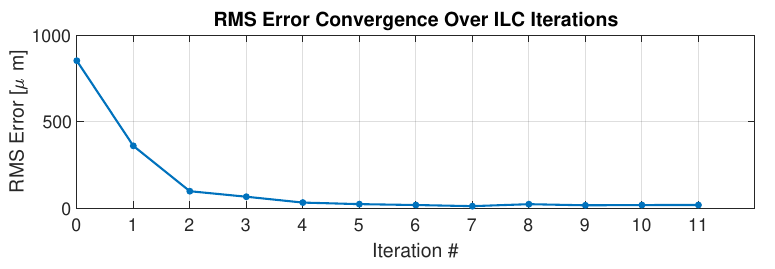}
    \caption{Convergence of the position RMS error over multiple iterations of ILC. The error stabilizes at $\approx25\mu m$ after 4 iterations}
    \label{fig:ILC_kinematic_convergence}
\end{figure}

To find the trajectory that would reduce tooltip deviations, the ILC framework described in \ref{subsec:ILC_kinematics} is utilized. This section focuses on the specific setup and implementation of the method for our injection task.\\
To begin, the range of motion desired for the rotation should be selected. While a full $360^\circ$ rotation can be used, a motion of $200^\circ$ was sufficient to achieve the desired twisting motion in our experiments. This range is divided into $10^\circ$ increments, resulting in a set of discrete tool orientations at which measurements are taken and joint values are updated during the ILC process. 
Additionally, the insertion range that will be used during surgery must be taken into account. As this experiment is to be done on the retina, a maximum insertion of $2mm$ is deemed appropriate. This range is split into $0.5mm$ increments, with calibration being performed for each value.\\
At each increment, OCT volume scans are captured to measure the position of the tooltip. The tooltip position is extracted using a combination of traditional image processing techniques. The volume scan is first binarized and, after noise removal, converted to a point cloud. Principle Component Analysis (PCA) is then performed on the shaft to find an approximate centerline for the tool, and a cylinder is fitted to the point cloud, with outliers being removed. The farthest point on the point cloud is then projected onto the centerline, and this location is selected as the tooltip location. The segmented tool can be seen in Fig. \ref{fig:tool_segmentation}. The positional error at each increment is computed relative to the tooltip position at the initial orientation, which is treated as the reference position we wish to maintain.\\
Next, the learning gain, which we defined as the inverse of the Jacobian matrix, must be calculated. The Jacobian is defined as the partial derivative of the positions with respect to each joint.
Because finding the Jacobian analytically is difficult, we choose a numerical approximation of the gradient at each orientation. This is done by perturbing each joint value slightly in the positive and negative directions while keeping the other joints constant. The partial derivative is then approximated as:
\begin{equation}
    \frac{\partial \mathbf{FK}(\mathbf{q})}{\partial q_i} \approx \frac{\mathbf{FK}(\mathbf{q} + \Delta q_i) - \mathbf{FK}(\mathbf{q} - \Delta q_i)}{2 \Delta q_i}.
\end{equation}
where $\mathbf{FK}$ denotes the forward kinematics of the robot, resulting in the X, Y, and Z positions of the tooltip at each orientation. Repeating this process for all joints will construct the Jacobian matrix. Because our goal is to command the 4th joint and find the required orientations of the remaining three, the Jacobian is only calculated for the first three joints. Note that due to the variety of unmodeled kinematic errors that we are trying to address, this forward kinematics relation is only an approximation and will add unknown errors to the output. This error will be mitigated through multiple iterations of ILC. \\
With the Jacobian know, the new joint positions can be calculated based on Eq. \ref{eq:ILC_kinematic_dq_update} and \ref{eq:ILC_kinematic_q_update}. Multiple iterations of this loop are first performed, using the nominal forward kinematics as feedback, until the error converges. This process constitutes one iteration of the ILC, and the outputs are applied to update the robot's joint values for the next iteration. The process continues until the tooltip error converges to within a chosen threshold or stabilizes across successive iterations. Fig. \ref{fig:ILC_kinematic_convergence} shows the RMS error convergencefor one representative trial, where the tooltip error was reduced from an initial $850\mu m$ to approximately $25\mu m$ within 4 iterations, with the following iterations resulting in negligible error change. Across repeated experiments with different initial conditions, the algorithm consistently converged to below $30\mu m$ error within 4-5 iterations, even when the starting error ranged from $200\mu m$ to over $1000\mu m$. This robustness demonstrates that the kinematic loop can reliably compensate for misalignments and calibration errors under varying conditions\\
The final updated joint values at each increment are then interpolated through bilinear interpolation over the insertion value and the rotation angle to create a smooth, continuous trajectory for the tool motion during simultaneous insertion and rotation. 

% \begin{figure}
%     \centering
%     \includegraphics[width=\linewidth]{Figures/ILC_Insertion_and_Rotation.eps}
%     \caption{Trajectory designed from interpolation of the outputs of the kinematic ILC. Joint 4 follows a sinusoidal rotation and Joint 3 moves in $2mm$ while all joints also rotate to compensate for errors. \hl{show only the one with insertion, add a line showing initial value, make sure it matches fig 10}}
%     \label{fig:fast_traj_sin}
% \end{figure}
\begin{figure*}[h]
    \centering
\includegraphics[width=0.8\linewidth]{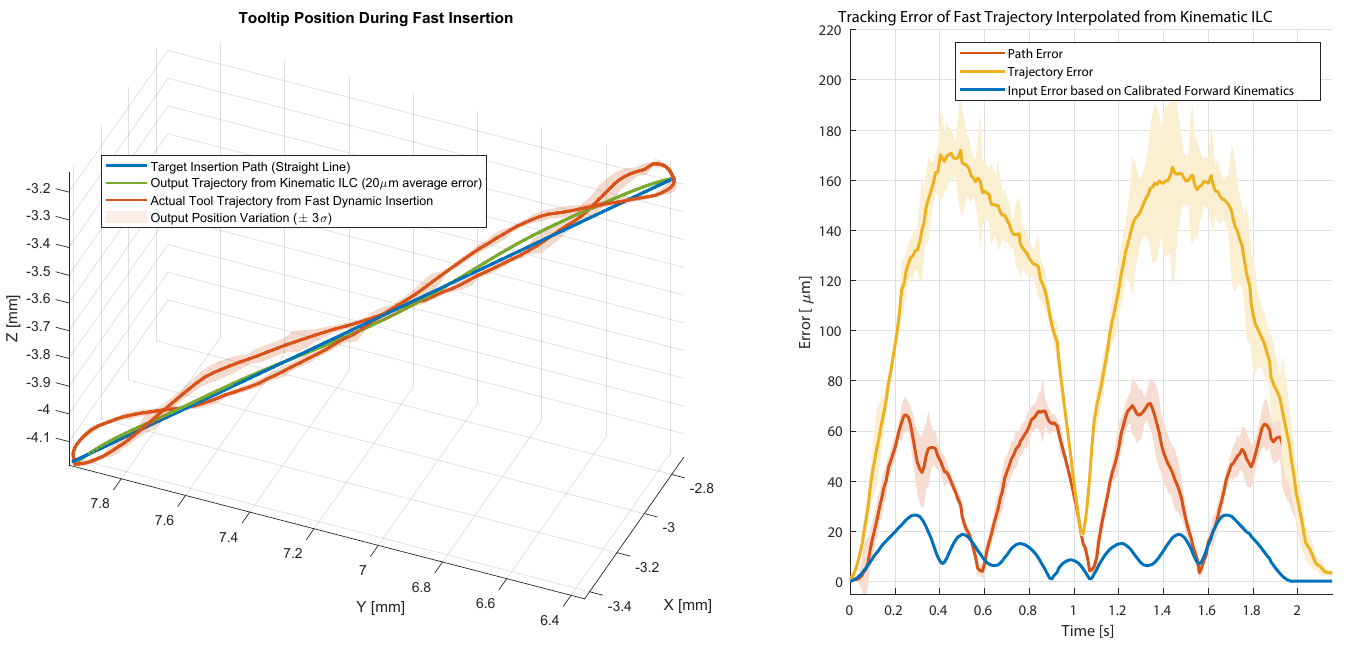}
    \caption{Tooltip positioning when performing interpolated fast trajectory over 30 iterations. Tracking error (red) compared to the nominal position (green) shows the necessity of a second step dynamics ILC.}
    \label{fig:fast_traj_tracking}
\end{figure*}
\subsubsection{Dynamics Loop Implementation}
\begin{figure*}
    \centering
    \includegraphics[width=\linewidth]{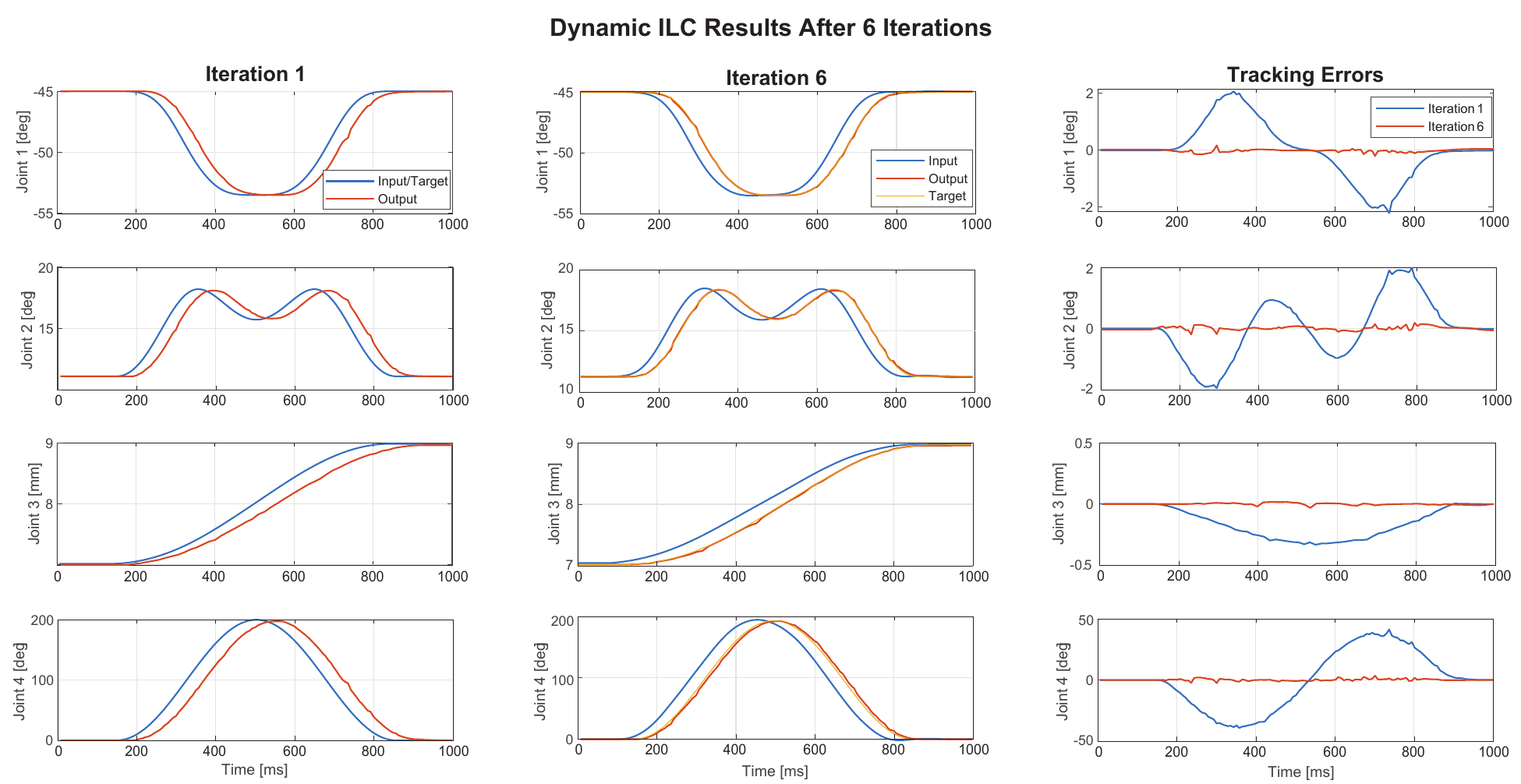}
    \caption{Sample output joint commands and tracking error of dynamic ILC after 6 iterations.}
    \label{fig:dynamic_ilc_results}
\end{figure*}

The trajectory determined after the kinematics loop is designed to be a sinusoidal cyclic rotation, which is continuously executed while the main surgical insertion is performed. 
% as seen in Fig. \ref{fig:dynamic_ilc_results}, Iteration 1. 
When the robot attempts to run this fast trajectory, actuator dynamics can result in actual joint outputs deviating from the commanded inputs. Figure \ref{fig:fast_traj_tracking} illustrates these effects, showing the trajectory and path errors, along with the tooltip path (red) resulting from directly implementing the interpolated trajectory. These positions are approximations based on the robot's forward kinematics, as the exact tooltip location cannot be tracked through imaging in the fast trajectory implementation. As such, it has a precision error of up to $20\mu m$, corresponding to the kinematics calibration error in this trial when following the reference trajectory (Fig. \ref{fig:dynamic_ilc_results}, Iteration 1), and is used as an approximate representation of tracking errors only. The "path error" represents the minimum distance to the closest point on the desired path, effectively ignoring pure delay effects, whereas the "trajectory error" compares each point with the corresponding target along time. To address the tracking error, a second stage ILC is implemented to compensate for the dynamics and ensure accurate trajectory tracking.\\
The first step in the dynamics correction process is to identify the transfer function for each joint, determining the relationship between joint input commands and the resulting outputs. This transfer function can then be inverted to be used as the learning gain in the ILC loop.\\
%This is done numerically through FIR inverse filter construction as explained in \cite{teng2015comparison}. 
Using the step response $g(n)$ measured for each joint, the impulse response coefficients $h(n)$ and Finite Impulse Response (FIR) representation $H(z)$ can be found:
\begin{gather}
    h(n) = \Delta g(n)=g(n)-g(n-1)\\
    H(z) = \sum_{k=0}^{N-1} h(k)z^{-k}
\end{gather}
This response is cropped, and a Blackman-Harris window function is applied for smoothing.
Preliminary experiments, as seen in the \textit{Iteration 1 Input/Output} graphs of Figure \ref{fig:dynamic_ilc_results}, show that the tracking errors arise primarily due to robot delay. Full FIR inversion was avoided because exact inversion amplifies noise and becomes unstable in the presence of non-minimum phase dynamics, actuator saturation, or model mismatch. Instead, we use phase-only inversion, which corrects the dominant delay-induced errors while leaving magnitude discrepancies to be corrected through iterative learning. The resulting filter is:
% , and as a result, we choose to perform only phase inversion rather than full FIR inverse filter design, resulting in:\\

\begin{equation}
    F(z) = H(z^{-1}) = \sum_{k=0}^{N-1}h(N-1-k)z^{-k}
\end{equation}
Where $F(z)$ denotes the learning gain for the dynamics ILC loop. This process is equivalent to reversing the order of the FIR coefficients $h(n)$, which allows for a straightforward implementation.
%The result then goes through Fast Fourier Transform to obtain the frequency domain response $H_w(n)$. Before inverting the frequency domain coefficients, a set of frequency domain low pass filter gains $Q(n)$ are designed. These gains are multiplied by the inverse of each coefficient in order to avoid numerical issues that will arise from dividing complex numbers close to zero. The inverse dynamic model can then be found by:\\
% \begin{equation}
%     F_{fi}(n) = (Q(0), ..., Q(N-1)) ./ (H_w(0), ..., H_w(N-1))
% \end{equation}
% where $./$ denotes element-wise division. The inverse Fourier transform can then provide the finite impulse response, and the inversion filter can be written as:
% \begin{equation}
%     F_{fi}(z^{-1}) = \sum_{k=0}^n f_{fi}(k)z^{-k} 
% \end{equation}
With these formulations, the input joint commands can be updated using Eq. \ref{eq:ILC_dynamics_update}.\\
After 6 iterations of dynamics ILC, the tracking error was substantially reduced, as seen Fig. \ref{fig:dynamic_ilc_results}. For joints 1–3, initial deviations of up to $2^\circ$ decreased to less than $0.1^\circ$, while in joint 4 the deviation dropped from nearly $50^\circ$ to about $1^\circ$. This represents more than an order-of-magnitude improvement in trajectory accuracy.
Residual error is primarily constrained by actuator resolution and speed, with performance bounded by saturation limits. Once saturation is reached, the system behavior becomes nonlinear and the identified transfer functions are no longer valid. Therefore, the trajectory must be designed carefully to remain within the actuators’ physical limits, but within this range, the dynamics loop enables accurate execution of cyclic rotation at surgical speeds.

% \begin{figure}
%     \centering
%     \includegraphics[width=0.5\linewidth]{Figures/L4_out_vs_in.png}
%     \caption{Graph of the input commands required to produce the desired output trajectory for joint 4 of the IRISS robot, computed through 4 iterations of the dynamic ILC.}
%     \label{fig:ILC_dynamics_L4_out_vs_in}
% \end{figure}

% \begin{figure}
%     \centering
%     \includegraphics[width=0.5\linewidth]{ILC_dynamics_tracking_errors.png}
%     \caption{Tracking error of each joint compared to the target (a) before ILC and (b) after 4 iterations of ILC.}
%     \label{fig:ILC_dynamics_tracking_errors}
% \end{figure}

\subsection{Surgical Evaluation}
With the rotation trajectory developed, it is then used during needle insertion when performing subretinal injection, and compared with results of similar experiments without rotation. The procedure is performed on ex-vivo pig eyes, using a 25/48g MedOne nanocannula. To simplify the problem and provide a reliable comparison, the tool is aligned with the target location, such that the injection can be performed by simply moving the needle forward within the selected $2mm$ range. This removes discrepancies and variations that can arise due to angle changes.\\
For each injection, the target depth is  chosen on the OCT image by a surgeon, and the robot is commanded to move to the target depth. Multiple insertions are performed on each eye, switching between direct insertion, and the developed bidirectional rotation trajectory. The insertion will continue until the surgeon confirms that the needle has reached the target, at which point, dye is injected, aiming to create a bleb in the subretinal space.\\
\begin{figure}
    \centering
    \includegraphics[width=\linewidth]{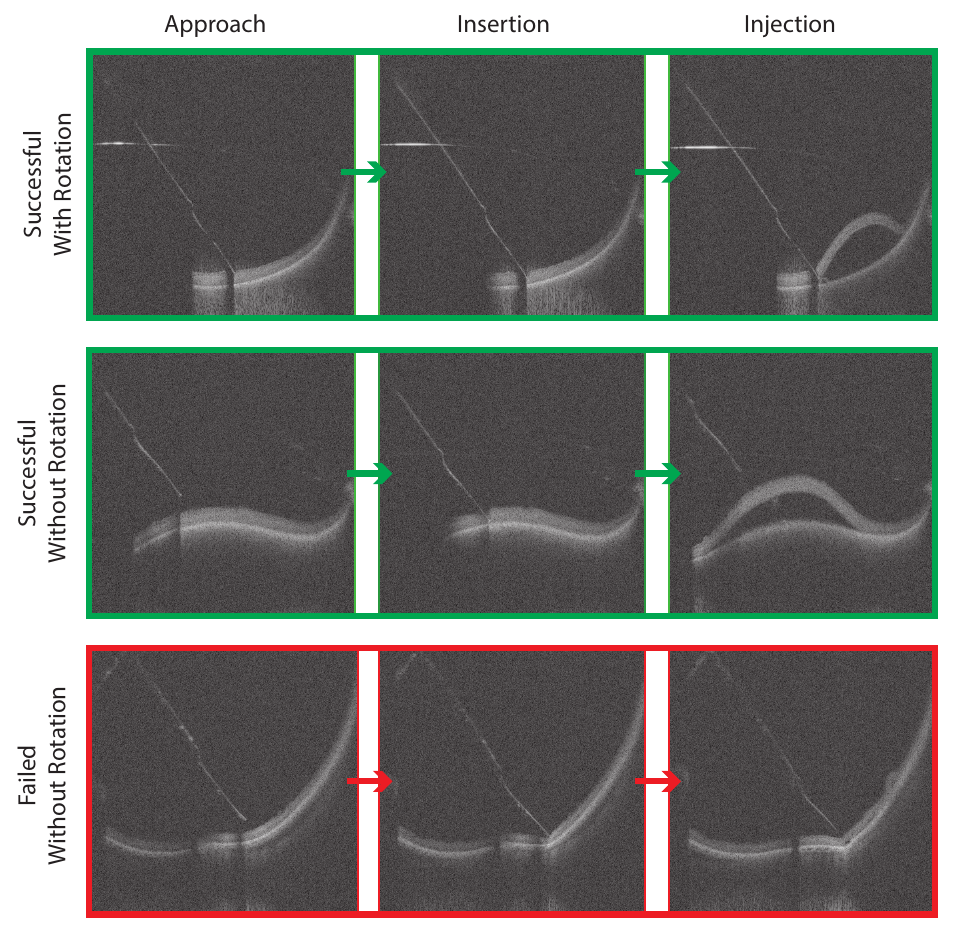}
    \caption{Subretinal injection results showing successful and failed needle insertion and injections.}
    \label{fig:surgical_results}
\end{figure}
Sample injection results can be seen in Fig. \ref{fig:surgical_results}, showing the tool approach, insertion into the retina, and resulting injection. In successful trials, the needle is inserted up to the RPE layer, and a bleb forms when injection is performed. However, in some trials without tool rotation, the tool failed to penetrate the retina, instead pushing and deforming the tissue. In these cases, the needle deformed the choroid, but did not break the surface of the retina to reach the RPE layer. As such, the trial was deemed unsuccessful, with no fluid being transferred to the subretinal space and no bleb was formed.\\
The experiments were performed on 4 pig eyes, with a total of 10 injections for each method. The results are summarized in Table \ref{tab:results}. The outcomes are split into: 1) Successful (Succ.), meaning acceptable injection and bleb formation as expected in the first attempt; 2) Failed (Fail.), when the needle was completely unable to penetrate the tissue and no bleb formed; 3) Delayed Success (Delay.), when initial penetration failed but eventual penetration and bleb formation occurred after perturbations, repeated insertions, or fluid effects. 
% It can be seen that with rotation, all trials were successful, with only one requiring slight adjustments before achieving correct bleb formation. Without rotation, however, $30\%$ of trials completely failed, and $20\%$ required repeated insertions and perturbation to break through the tissue. 
Overall, rotational insertion achieved a $90\%$ immediate success rate (9/10), compared to $50\%$ (5/10) for straight insertion. Failures dropped from 3 to 0, and delayed successes decreased from 2 to 1. These improved outcomes were only possible because the ILC framework constrained tooltip motion during rotation to within $\approx20\mu m$, keeping the trajectory safely inside the retinal thickness. Without this compensation, rotational insertion would have caused large tooltip excursions and tissue tearing, making the maneuver infeasible. This demonstrates that rotational insertion, when stabilized by ILC, not only improves reliability but also reduces the need for repeated attempts, which can cause greater tissue trauma.
\begin{table}[ht]
\centering
\begin{tabular}{|l|c|c|c|c|c|c|}
\hline
\textbf{Eye Num.} & \multicolumn{3}{c|}{\textbf{With Rotation}} & \multicolumn{3}{c|}{\textbf{Without Rotation}} \\
\hline
 & \textbf{Succ.} & \textbf{Fail.} & \textbf{Delay.} & \textbf{Succ.} & \textbf{Fail.} & \textbf{Delay.} \\
\hline
0 & 2 & 0 & 0 & 1 & 2 & 0 \\
1 & 1 & 0 & 1 & 0 & 0 & 1 \\
2 & 3 & 0 & 0 & 2 & 0 & 1 \\
3 & 3 & 0 & 0 & 2 & 1 & 0 \\
\hline
\textbf{Total} & \textbf{9} & \textbf{0} & \textbf{1} & \textbf{5} & \textbf{3} & \textbf{2} \\
\hline
\end{tabular}
\caption{Experiment outcomes for subretinal injection with and without rotation. Values represent number of successful, failed, and delayed successes.}
\label{tab:results}
\end{table}

%A similar result was seen during vein cannulation, with pure insertion requiring assistants to pull the vein tight in order to enable cannulation, but simply clasping the vein was sufficient to perform cannulation with rotation.\\

\section{Conclusion}
The results demonstrate that integrating rotational tool motion, enabled through ILC based trajectory design, significantly improves the reliability and precision of the tool rotation, allowing it to be performed during surgery within the required tolerances. This was not previously possible due to positioning errors that well exceeded retinal thickness and allowable surgical limits.
With the feasibility of the rotation confirmed, the strategy was then tested in the context of subretinal injection to assess its effects on surgery. After rotation, success rates increased from 5 out of 10 trials without rotation, to 9 out of 10 trials with rotation. Notably, the number of trials where the needle completely failed to penetrate the retina dropped from 3 to zero, while the number of trials requiring slight adjustment to allow successful injection decreased from 3 to 1, showing not only higher success rates, but also more consistent and predictable outcomes.\\
This improvement is attributed to two key factors. First, the rotational motion itself reduces the resistance at the tool-tissue interface, allowing smoother penetration. Second and more critically, the ILC framework allows this rotation to happen without the tooltip moving, something that would result in retinal tearing and a complete failure to keep fluid in the retinal space, as the large tear would immediately leak.\\
Importantly, this approach does not require precise modeling of the robot or tissue interaction. Instead, it uses measured error signals from previous trials to incrementally adapt control inputs, making this data driven approach adaptable to different procedures. While the current implementation focuses on subretinal injection, the framework can be extended to other procedures requiring precise insertions, such as vein cannulation and biopsy.\\
Despite the shown benefits, several limitations remain and will guide future work.
One issue is that the dynamic correction achievable through the current ILC framework is constrained by the physical limits of the actuators. Without a mechanism to detect actuator saturation, there is a risk of the model attempting to overcorrect beyond the system’s capabilities, leading to degraded performance. Future work will incorporate models or feedback mechanisms that can detect such instances and prevent overcompensation during iterative updates.\\
In addition, the current reliance on relatively slow OCT volumetric scans makes trajectory planning time-consuming and limited to a specific orientation at each injection site. With the development and integration of faster swept-source OCT systems, it is possible to generate full 3D maps of target commands over the whole surgical workspace. This would enable the rotation axis and insertion trajectory to be optimized at any location across the retina, significantly expanding the flexibility and robustness of the method.\\
Another limitation is the reliance on manual alignment and setup. Further automation and feedback is needed to reduce reliance on manual alignment and avoid human errors. Improvements in initial tool registration and implementing real-time tissue feedback and trajectory correction are critical steps toward enabling fully automated, precision-guided injections across clinical scenarios.\\
Finally, while the immediate experimental outcomes showed clear improvements in penetration and bleb formation, the long term effects on the retina, including potential tissue trauma from rotation, remain unknown. Future studies must transition to performing the procedure in closed-eye models and ultimately in-vivo experiments, followed by histological analysis to assess tissue integrity, bleb formation, and retinal quality after surgery.\\

Overall, this work establishes a foundation for integrating precision bidirectional rotation into delicate surgical tasks where positional accuracy is critical. By combining data-driven trajectory design with high-precision robotic motion and rotation for needle penetration, the proposed method enhances the safety, reliability, and effectiveness of subretinal injection, and has the potential to improve a broad range of surgical procedures requiring accurate needle penetration.

\bibliographystyle{IEEEtran}
\bibliography{references}

% Generated by IEEEtran.bst, version: 1.14 (2015/08/26)
\begin{thebibliography}{10}
\providecommand{\url}[1]{#1}
\csname url@samestyle\endcsname
\providecommand{\newblock}{\relax}
\providecommand{\bibinfo}[2]{#2}
\providecommand{\BIBentrySTDinterwordspacing}{\spaceskip=0pt\relax}
\providecommand{\BIBentryALTinterwordstretchfactor}{4}
\providecommand{\BIBentryALTinterwordspacing}{\spaceskip=\fontdimen2\font plus
\BIBentryALTinterwordstretchfactor\fontdimen3\font minus \fontdimen4\font\relax}
\providecommand{\BIBforeignlanguage}[2]{{%
\expandafter\ifx\csname l@#1\endcsname\relax
\typeout{** WARNING: IEEEtran.bst: No hyphenation pattern has been}%
\typeout{** loaded for the language `#1'. Using the pattern for}%
\typeout{** the default language instead.}%
\else
\language=\csname l@#1\endcsname
\fi
#2}}
\providecommand{\BIBdecl}{\relax}
\BIBdecl

\bibitem{iwasaki2012high}
M.~Iwasaki, K.~Seki, and Y.~Maeda, ``High-precision motion control techniques: A promising approach to improving motion performance,'' \emph{IEEE Industrial Electronics Magazine}, vol.~6, no.~1, pp. 32--40, 2012.

\bibitem{singh2002physiological}
S.~Singh and C.~Riviere, ``Physiological tremor amplitude during retinal microsurgery,'' in \emph{Proceedings of the IEEE 28th Annual Northeast Bioengineering Conference (IEEE Cat. No. 02CH37342)}.\hskip 1em plus 0.5em minus 0.4em\relax IEEE, 2002, pp. 171--172.

\bibitem{iordachita2022robotic}
I.~I. Iordachita, M.~D. De~Smet, G.~Naus, M.~Mitsuishi, and C.~N. Riviere, ``Robotic assistance for intraocular microsurgery: Challenges and perspectives,'' \emph{Proceedings of the IEEE}, vol. 110, no.~7, pp. 893--908, 2022.

\bibitem{tsumura2019needle}
R.~Tsumura, Y.~Takishita, and H.~Iwata, ``Needle insertion control method for minimizing both deflection and tissue damage,'' \emph{Journal of Medical Robotics Research}, vol.~4, no.~01, p. 1842005, 2019.

\bibitem{pannek2024exploring}
S.~Pannek, S.~Dehghani, M.~Sommersperger, P.~Zhang, P.~Gehlbach, M.~A. Nasseri, I.~Iordachita, and N.~Navab, ``Exploring the needle tip interaction force with retinal tissue deformation in vitreoretinal surgery,'' in \emph{2024 IEEE International Conference on Robotics and Automation (ICRA)}.\hskip 1em plus 0.5em minus 0.4em\relax IEEE, 2024, pp. 16\,999--17\,005.

\bibitem{hochman2001vitro}
M.~N. Hochman and M.~J. Friedman, ``An in vitro study of needle force penetration comparing a standard linear insertion to the new bidirectional rotation insertion technique.'' \emph{Quintessence International}, vol.~32, no.~10, 2001.

\bibitem{abolhassani2006control}
N.~Abolhassani, R.~Patel, and M.~Moallem, ``Control of soft tissue deformation during robotic needle insertion,'' \emph{Minimally invasive therapy \& allied technologies}, vol.~15, no.~3, pp. 165--176, 2006.

\bibitem{tsumura2016histological}
R.~Tsumura, Y.~Takishita, Y.~Fukushima, and H.~Iwata, ``Histological evaluation of tissue damage caused by rotational needle insertion,'' in \emph{2016 38th annual international conference of the IEEE engineering in medicine and biology society (EMBC)}.\hskip 1em plus 0.5em minus 0.4em\relax IEEE, 2016, pp. 5120--5123.

\bibitem{lin2020simultaneously}
C.-L. Lin and Y.-A. Huang, ``Simultaneously reducing cutting force and tissue damage in needle insertion with rotation,'' \emph{IEEE Transactions on Biomedical Engineering}, vol.~67, no.~11, pp. 3195--3202, 2020.

\bibitem{millington2019innovative}
J.~Millington, R.~P. Monfared, and D.~Vera, ``Innovative mechanism to identify robot alignment in an automation system,'' \emph{Robotics and Autonomous Systems}, vol. 114, pp. 144--154, 2019.

\bibitem{wilson2018intraocular}
J.~T. Wilson, M.~J. Gerber, S.~W. Prince, C.-W. Chen, S.~D. Schwartz, J.-P. Hubschman, and T.-C. Tsao, ``Intraocular robotic interventional surgical system (iriss): Mechanical design, evaluation, and master--slave manipulation,'' \emph{The International Journal of Medical Robotics and Computer Assisted Surgery}, vol.~14, no.~1, p. e1842, 2018.

\bibitem{kim2015effects}
L.~H. Kim, C.~Bargar, Y.~Che, and A.~M. Okamura, ``Effects of master-slave tool misalignment in a teleoperated surgical robot,'' in \emph{2015 IEEE International Conference on Robotics and Automation (ICRA)}.\hskip 1em plus 0.5em minus 0.4em\relax IEEE, 2015, pp. 5364--5370.

\bibitem{beasley2004kinematic}
R.~A. Beasley, R.~D. Howe, and P.~E. Dupont, ``Kinematic error correction for minimally invasive surgical robots,'' in \emph{IEEE International Conference on Robotics and Automation, 2004. Proceedings. ICRA'04. 2004}, vol.~1.\hskip 1em plus 0.5em minus 0.4em\relax IEEE, 2004, pp. 358--364.

\bibitem{bai2021error}
M.~Bai, M.~Zhang, H.~Zhang, L.~Pang, J.~Zhao, and C.~Gao, ``An error compensation method for surgical robot based on rcm mechanism,'' \emph{IEEE Access}, vol.~9, pp. 140\,747--140\,758, 2021.

\bibitem{huang2024operational}
F.~Huang, H.~Sang, F.~Liu, and R.~Han, ``Operational space robust impedance control of the redundant surgical robot for minimally invasive surgery,'' \emph{Medical \& Biological Engineering \& Computing}, pp. 1--15, 2024.

\bibitem{cursi2020model}
F.~Cursi, V.~Modugno, and P.~Kormushev, ``Model predictive control for a tendon-driven surgical robot with safety constraints in kinematics and dynamics,'' in \emph{2020 IEEE/RSJ International Conference on Intelligent Robots and Systems (IROS)}.\hskip 1em plus 0.5em minus 0.4em\relax IEEE, 2020, pp. 7653--7660.

\bibitem{ogihara2024intraoperative}
A.~Ogihara, M.~Omata, H.~Shidei, S.~Mitsuboshi, H.~Aoshima, T.~Isaka, T.~Matsumoto, and M.~Kanzaki, ``Intraoperative robotic surgical system-related problems in robot-assisted thoracoscopic surgery,'' \emph{General Thoracic and Cardiovascular Surgery}, pp. 1--6, 2024.

\bibitem{omisore2020motion}
O.~M. Omisore, S.~Han, Y.~Al-Handarish, W.~Du, W.~Duan, T.~O. Akinyemi, and L.~Wang, ``Motion and trajectory constraints control modeling for flexible surgical robotic systems,'' \emph{Micromachines}, vol.~11, no.~4, p. 386, 2020.

\bibitem{li2018needle}
P.~Li, Z.~Yang, and S.~Jiang, ``Needle-tissue interactive mechanism and steering control in image-guided robot-assisted minimally invasive surgery: a review,'' \emph{Medical \& Biological Engineering \& Computing}, vol.~56, pp. 931--949, 2018.

\bibitem{spong2020robot}
M.~W. Spong, S.~Hutchinson, and M.~Vidyasagar, \emph{Robot modeling and control}.\hskip 1em plus 0.5em minus 0.4em\relax John Wiley \& Sons, 2020.

\bibitem{wei2018adaptive}
B.~Wei, ``Adaptive control design and stability analysis of robotic manipulators,'' in \emph{Actuators}, vol.~7, no.~4.\hskip 1em plus 0.5em minus 0.4em\relax MDPI, 2018, p.~89.

\bibitem{kaneko1997repetitive}
K.~Kaneko and R.~Horowitz, ``Repetitive and adaptive control of robot manipulators with velocity estimation,'' \emph{IEEE Transactions on Robotics and Automation}, vol.~13, no.~2, pp. 204--217, 1997.

\bibitem{abtahi2023automatic}
M.~Abtahi, M.~Rabbani, and S.~Nazari, ``An automatic tuning mpc with application to ecological cruise control,'' \emph{IFAC-PapersOnLine}, vol.~56, no.~3, pp. 265--270, 2023.

\bibitem{kim1996iterative}
D.-I. Kim and S.~Kim, ``An iterative learning control method with application for cnc machine tools,'' \emph{IEEE Transactions on Industry Applications}, vol.~32, no.~1, pp. 66--72, 1996.

\bibitem{hagqvist2015resistance}
P.~Hagqvist, A.~Herali{\'c}, A.-K. Christiansson, and B.~Lennartson, ``Resistance based iterative learning control of additive manufacturing with wire,'' \emph{Mechatronics}, vol.~31, pp. 116--123, 2015.

\bibitem{blanken2016design}
L.~Blanken, J.~Willems, S.~Koekebakker, and T.~Oomen, ``Design techniques for multivariable ilc: Application to an industrial flatbed printer,'' \emph{IFAC-PapersOnLine}, vol.~49, no.~21, pp. 213--221, 2016.

\bibitem{dijkstra2002convergence}
B.~G. Dijkstra and O.~H. Bosgra, ``Convergence design considerations of low order q-ilc for closed loop systems, implemented on a high precision wafer stage,'' in \emph{Proceedings of the 41st IEEE Conference on Decision and Control, 2002.}, vol.~3.\hskip 1em plus 0.5em minus 0.4em\relax IEEE, 2002, pp. 2494--2499.

\bibitem{lee2021industrial}
Y.-H. Lee, S.-C. Hsu, T.-Y. Chi, Y.-Y. Du, J.-S. Hu, and T.-C. Tsao, ``Industrial robot accurate trajectory generation by nested loop iterative learning control,'' \emph{Mechatronics}, vol.~74, p. 102487, 2021.

\bibitem{longman2000iterative}
R.~W. Longman, ``Iterative learning control and repetitive control for engineering practice,'' \emph{International journal of control}, vol.~73, no.~10, pp. 930--954, 2000.

\bibitem{norrlof2002time}
M.~Norrl{\"o}f and S.~Gunnarsson, ``Time and frequency domain convergence properties in iterative learning control,'' \emph{International Journal of Control}, vol.~75, no.~14, pp. 1114--1126, 2002.

\bibitem{gunnarsson2001design}
S.~Gunnarsson and M.~Norrl{\"o}f, ``On the design of ilc algorithms using optimization,'' \emph{Automatica}, vol.~37, no.~12, pp. 2011--2016, 2001.

\bibitem{tao2021robust}
H.~Tao, X.~Li, W.~Paszke, V.~Stojanovic, and H.~Yang, ``Robust pd-type iterative learning control for discrete systems with multiple time-delays subjected to polytopic uncertainty and restricted frequency-domain,'' \emph{Multidimensional Systems and Signal Processing}, vol.~32, no.~2, pp. 671--692, 2021.

\bibitem{wang2015nonparametric}
C.~Wang, Y.~Zhao, Y.~Chen, and M.~Tomizuka, ``Nonparametric statistical learning control of robot manipulators for trajectory or contour tracking,'' \emph{Robotics and Computer-Integrated Manufacturing}, vol.~35, pp. 96--103, 2015.

\bibitem{lee2022data}
Y.-H. Lee, S.~Rai, and T.-C. Tsao, ``Data-driven iterative learning control of nonlinear systems by adaptive model matching,'' \emph{IEEE/ASME Transactions on Mechatronics}, vol.~27, no.~6, pp. 5626--5636, 2022.

\bibitem{chen2018intraocular}
C.-W. Chen, Y.-H. Lee, M.~J. Gerber, H.~Cheng, Y.-C. Yang, A.~Govetto, A.~A. Francone, S.~Soatto, W.~S. Grundfest, J.-P. Hubschman \emph{et~al.}, ``Intraocular robotic interventional surgical system (iriss): semi-automated oct-guided cataract removal,'' \emph{The International Journal of Medical Robotics and Computer Assisted Surgery}, vol.~14, no.~6, p. e1949, 2018.

\bibitem{lee2023accurate}
Y.-H. Lee, Y.-T. Lai, M.~J. Gerber, J.~Dodds, J.-P. Hubschman, J.~Rosen, and T.-C. Tsao, ``Accurate robotic posterior capsule polishing with tissue stabilization,'' \emph{IEEE/ASME Transactions on Mechatronics}, vol.~29, no.~1, pp. 290--300, 2023.

\bibitem{gerber2021robotic}
M.~J. Gerber, J.-P. Hubschman, and T.-C. Tsao, ``Robotic posterior capsule polishing by optical coherence tomography image guidance,'' \emph{The International Journal of Medical Robotics and Computer Assisted Surgery}, vol.~17, no.~3, p. e2248, 2021.

\bibitem{chen2019semiautomated}
C.-W. Chen, A.~A. Francone, M.~J. Gerber, Y.-H. Lee, A.~Govetto, T.-C. Tsao, and J.-P. Hubschman, ``Semiautomated optical coherence tomography-guided robotic surgery for porcine lens removal,'' \emph{Journal of Cataract \& Refractive Surgery}, vol.~45, no.~11, pp. 1665--1669, 2019.

\bibitem{barzelay2024transforming}
A.~Barzelay, Y.-T. Lai, M.~Reyes, S.~Schwartz, T.-C. Tsao \emph{et~al.}, ``Transforming retinal surgery: Pioneering robotic arm and intraocular imaging probes redefine the landscape of retinal surgery,'' \emph{Investigative Ophthalmology \& Visual Science}, vol.~65, no.~7, pp. 3860--3860, 2024.

\bibitem{gerber2020automated}
M.~J. Gerber, J.-P. Hubschman, and T.-C. Tsao, ``Automated retinal vein cannulation on silicone phantoms using optical-coherence-tomography-guided robotic manipulations,'' \emph{IEEE/ASME Transactions on Mechatronics}, vol.~26, no.~5, pp. 2758--2769, 2020.

\bibitem{lai2025safe}
Y.-T. Lai, Y.~Foroutani, A.~Barzelay, and T.-C. Tsao, ``Safe robotic capsule cleaning with integrated transpupillary and intraocular optical coherence tomography,'' \emph{arXiv preprint arXiv:2507.13650}, 2025.

\bibitem{chang2012high}
H.~L. Chang and T.-C. Tsao, ``High-sampling rate dynamic inversion—filter realization and applications in digital control,'' \emph{IEEE/ASME Transactions on Mechatronics}, vol.~19, no.~1, pp. 238--248, 2012.

\bibitem{tsao1987adaptive}
T.-C. Tsao and M.~Tomizuka, ``Adaptive zero phase error tracking algorithm for digital control,'' 1987.

\bibitem{rigney2009nonminimum}
B.~P. Rigney, L.~Y. Pao, and D.~A. Lawrence, ``Nonminimum phase dynamic inversion for settle time applications,'' \emph{IEEE Transactions on Control Systems Technology}, vol.~17, no.~5, pp. 989--1005, 2009.

\bibitem{kim2008model}
K.-S. Kim and Q.~Zou, ``Model-less inversion-based iterative control for output tracking: piezo actuator example,'' in \emph{2008 American Control Conference}.\hskip 1em plus 0.5em minus 0.4em\relax IEEE, 2008, pp. 2710--2715.

\bibitem{teng2015comparison}
K.-T. Teng and T.-C. Tsao, ``A comparison of inversion based iterative learning control algorithms,'' in \emph{2015 American control conference (ACC)}.\hskip 1em plus 0.5em minus 0.4em\relax IEEE, 2015, pp. 3564--3569.

\end{thebibliography}

\end{document}